\documentclass[10pt,twocolumn,letterpaper]{article}

\usepackage{cvpr}
\usepackage{times}
\usepackage{epsfig}
\usepackage{graphicx}
\usepackage{amsmath}
\usepackage{amssymb}
\usepackage{tabu}
\usepackage{multirow}

\usepackage[breaklinks=true,bookmarks=false]{hyperref}

\cvprfinalcopy 

\def\cvprPaperID{3687} 

\ifcvprfinal\fi
\begin{document}

\title{Robust Classification with Convolutional Prototype Learning}

\author{Hong-Ming Yang$^{1,2}$, Xu-Yao Zhang$^{1,2}$, Fei Yin$^{1,2}$, Cheng-Lin Liu$^{1,2,3}$\\
$^1$NLPR, Institute of Automation, Chinese Academy of Sciences, Beijing, P.R. China\\
$^2$University of Chinese Academy of Sciences, Beijing, P.R. China\\
$^3$CAS Center for Excellence of Brain Science and Intelligence Technology, Beijing, P.R. China\\
{\tt\small \{hongming.yang, xyz, fyin, liucl\}@nlpr.ia.ac.cn}
}

\maketitle
\thispagestyle{empty}

\begin{abstract}
   Convolutional neural networks (CNNs) have been widely used for image classification. Despite its high accuracies, CNN has been shown to be easily fooled by some adversarial examples, indicating that CNN is not robust enough for pattern classification. In this paper, we argue that the lack of robustness for CNN is caused by the softmax layer, which is a totally discriminative model and based on the assumption of closed world (i.e., with a fixed number of categories). To improve the robustness, we propose a novel learning framework called convolutional prototype learning (CPL). The advantage of using prototypes is that it can well handle the open world recognition problem and therefore improve the robustness. Under the framework of CPL, we design multiple classification criteria to train the network. Moreover, a prototype loss (PL) is proposed as a regularization to improve the intra-class compactness of the feature representation, which can be viewed as a generative model based on the Gaussian assumption of different classes. Experiments on several datasets demonstrate that CPL can achieve comparable or even better results than traditional CNN, and from the robustness perspective, CPL shows great advantages for both the rejection and incremental category learning tasks.
\end{abstract}

\vspace{-0.5cm}\section{Introduction}

In recent years, convolutional neural networks \cite{Cun1990Handwritten} (CNNs) have achieved great success for pattern recognition and computer vision, leading to important progress in a variety of tasks, like image classification \cite{Krizhevsky2012ImageNet, He2016Deep, he2015delving, simonyan2014very}, object detection \cite{dai2016r, ren2015faster, He_2017_ICCV}, instance segmentation \cite{pinheiro2015learning, he2017mask} and so on.

\begin{figure}
\begin{center}
\includegraphics[width=0.7\linewidth]{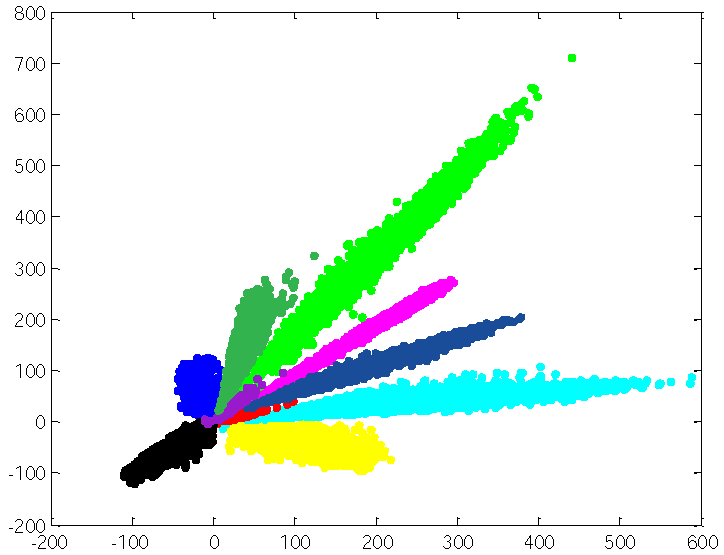}
\end{center}
   \caption{Feature representation learned by traditional CNN model on MNIST. Different colors represent different classes. It is shown that the inter-class variation is even smaller than the intra-class variation.}
\label{fig:softmax}
\vspace{-0.5cm}
\end{figure}

Despite the success of CNN, it still suffer from some serious problems. One example is the existence of adversarial samples \cite{szegedy2013intriguing}, when we add small noises or make some small changes to the initial samples, CNN will give different predictions for these samples with high confidence, although visually we can hardly find any significant changes in the images. Another example is the rejection ability of CNN, when feed a sample from an unseen class to CNN, it will still allocate the sample to a known class with high confidence. These two phenomena indicate that CNN is not robust, though it can achieve human-level or even better accuracy on some specific datasets, its performance will degenerate obviously in the complex scenes of the reality. This greatly limits the application of CNN in real worlds. The main reasons for these problems include two aspects: First, CNN is a purely discriminative model, it essentially learns a partition of the ``whole" feature space, therefore, the samples from unseen classes are still predicted to some specific regions under the partition, and CNN still views these samples as some known classes with high confidence. This explains why the rejection ability of CNN is poor; Second, from the perspective of representation learning, the learned representation of CNN is linear separable, see Fig. \ref{fig:softmax} for an illustration, and under this kind of representation, the inter-class distance is sometimes even smaller than the intra-class distance, this significantly reduces the robustness of CNN in real and complicated environments.

Several methods have been proposed to improve the robustness of CNN and most of them concentrate on designing better loss functions. \cite{Sun2014Deep} and \cite{Schroff2015FaceNet} proposed the contrastive loss and triplet loss to learn a more robust feature representation, in which the input pairs and triplets need to be carefully selected from the training data to ensure the convergence and stability. \cite{Wen2016A} proposed the center loss to improve the performance of softmax-based CNN, however, the centers can not be learned jointly with the CNN and are only updated according to some pre-defined rules rather than learned directly from data. Our CPL is more general than the center loss, since we totally abandon softmax layer and all the prototypes are learned automatically from data. Moreover, previous work of \cite{Liu2016Large} and \cite{Liu2017SphereFace} also made improvements and extensions for the softmax based loss but they still kept the softmax layer with the traditional framework of CNN for classification.

In this paper, we propose a novel framework called convolutional prototype learning (CPL) for image classification. In the bottom of CPL, the convolutional layers are used to extract discriminative features just like traditional CNN, but in the top of CPL we assign multiple prototypes to represent different classes. The classification is simply implemented by finding the nearest prototype (using Euclidean distance) in the feature space. We design multiple loss functions for this framework, making the CNN feature extractor and the prototypes being learned jointly from the raw data. Therefore, the whole framework can be trained efficiently and effectively. Experiments on several datasets demonstrate that the CPL framework can achieve comparable or even better classification accuracies compared with traditional CNN models.

Benefited from the prototype-based decision function, a natural prototype loss (PL) can be added to our CPL framework, to pull the feature vector closer to their corresponding prototypes (genuine class representation). The PL is akin to the maximum likelihood (ML) regularization proposed in \cite{liu2004effects, liu2004discriminative}. On one hand, it acts like a regularizer, which can prevents the model form over-fitting and improves the performance of CPL. On the other hand, it can also improves the intra-class compactness in feature representation. Therefore, the final learned representation is intra-class compact and inter-class separable, which makes the representation more discriminative and robust. From the perspective of probability, our CPL and PL framework essentially extract, transform, and model the data of each class as a Gaussian mixture distribution and the prototypes act as the means of Gaussian components for each class, this enables integrating probabilistic methods such as Bayesian models into our framework. Compared with the traditional CNN framework, we do not make partition for the ``whole'' feature space, but project the samples to some specific regions of the feature space (near the prototypes), thus our model is more robust to samples from unseen classes and more suitable for rejection. CPL can also be viewed as a hybrid discriminative and generative model (like the discriminative density model in \cite{liu2004effects}) which will lead to better generalization performance.

\section{Related works}

In this section, we describe related works from two aspects: robust representation learning and prototype learning. Most recent methods concentrate on modifying or proposing new loss functions to learn discriminative and robust representations. Among these methods, \cite{Sun2014Deep} combined the cross entropy loss and contrastive loss \cite{Hadsell2006Dimensionality} to train the CNN, the cross entropy loss can increase the inter-personal variations while the contrastive loss can reduce the intra-personal variations, and both losses guide the CNN to learn more discriminative representations. \cite{Schroff2015FaceNet} designed a triplet loss for CNN to learn representations in a compact Euclidean space where distances directly correspond to a measure of similarity, and the learned representation performs well on several tasks including recognition, verification, and clustering. \cite{Wen2016A} proposed a center loss and combine it with cross entropy loss to train the CNN for learning more discriminative features, and they also propose a mini-batch based update method for the centers, which was proved to be useful for face recognition and verification. \cite{Liu2016Large} proposed a generalized large-margin softmax loss which explicitly encourages intra-class compactness and inter-class separability between learned representations, making the representation more discriminative and robust. \cite{Liu2017SphereFace} further proposed a angular soft-max loss, which can ensure the learned representations more angularly discriminative, and this method was proved to be efficient under open-set protocols.

Prototype learning is a classical and representative method in patter recognition society. The earliest prototype learning method is k-nearest-neighbor (K-NN). In order to reduce the heavy burden of storages space and computation requirement of K-NN, an improvement called learning vector quantization (LVQ) is proposed \cite{Kohonen1990The}. The LVQ has been studied in many works and it has a lot of variations. According to the updating methods of the prototypes, we can classify the LVQ methods into two main categories. The first category concentrates on designing suitable updating conditions and rules to learn the prototypes, and the representing works include \cite{Kohonen1990The,Kohonen2012Improved,Geva1991Adaptive,Lee1994Optimal,Liu1997High}. The other category learns the prototypes in a parameter optimization way, by defining loss functions with regard to the prototypes and learning the prototypes through optimizing the loss functions. The representative methods include \cite{Sato1996Generalized,Sato1998A,Huang1996A,Miller1996A,Decaestecker1997Finding}. A detailed review and evaluation of the prototype based learning methods can be found in \cite{liu2001evaluation}. Previous prototype learning methods are mainly based on hand-designed features and they were widely used in different pattern recognition tasks before the arrival of CNN. To the best of our knowledge, this is the first work on combining the prototype based classifiers with deep convolutional neural networks for both high accuracy and robust pattern classification.

\section{Convolutional prototype learning}

\subsection{Architecture of the framework}\label{structure}
Compared with hand-designed features, the features automatically learned from data usually perform better for classification. Thus, we use a CNN as feature extractor in our framework, which is denoted as $f(x;\theta)$, $x$ and $\theta$ denote the raw input and parameters of the CNN respectively. Different from the traditional CNN which use softmax layer for linear classification on the learned features, we maintain and learn several prototypes on the features for each class and use prototype matching for classification.
The prototypes are denoted as $m_{ij}$ where $i\in\{1,2,...,C\}$ represents the index of the classes and $j\in\{1,2,...,K\}$ represents the index of the prototypes in each class. Here we assume each class having equal number of $K$ prototypes and this assumption can be easily relaxed in real application.

\begin{figure}
\begin{center}
\includegraphics[width=0.9\linewidth]{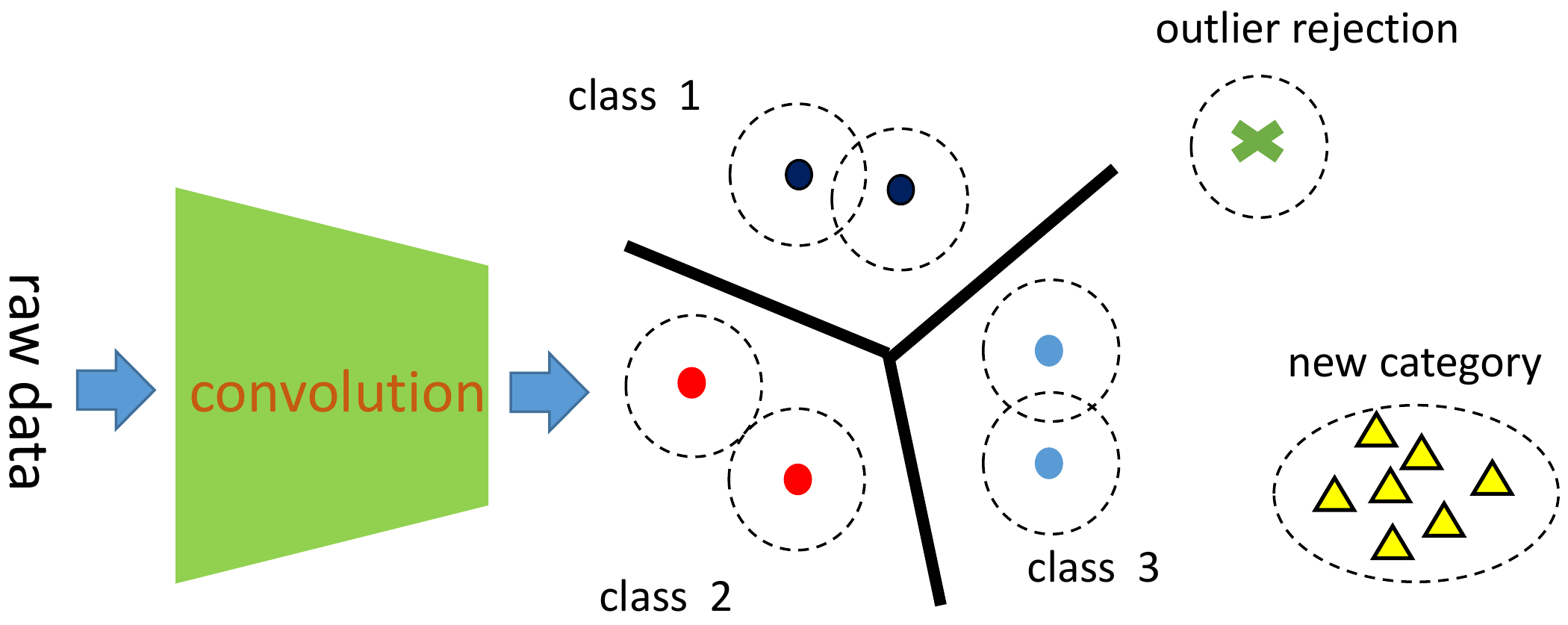}
\end{center}
   \caption{An illustration of convolutional prototype learning.}
\label{fig:structure}
\vspace{-0.3cm}
\end{figure}

The CNN feature extractor $f(x;\theta)$ and the prototypes $\{m_{ij}\}$ are jointly trained from data. In the classification stage, we classify the objects by prototype matching, i.e., we find the nearest prototype according the Euclidean distance and assign the class of this prototype to the particular object. A graphic description of our framework can be seen in Fig. \ref{fig:structure}.

\subsection{Feedforward for prediction}
Given an input pattern $x$, we first get its abstract representation by the CNN feature extractor, then we compare the abstract feature with all prototypes and classify it to the category where the nearest prototype belongs to:
\begin{equation} x \in class \ \ \arg\max_{i=1}^{C} g_{i}(x)\label{prediction} \end{equation}
where $g_{i}(x)$ is the discriminant function for class $i$:
\begin{equation} g_{i}(x) = -\min_{j=1}^{K} \left\| f(x;\theta) - m_{ij} \right\|_{2}^{2}\label{discriminant} \end{equation}

\subsection{Backward for training}\label{training}
The trainable parameters in our framework are composed by two parts. One is the parameters of the CNN extractor, which is denoted as $\theta$; the other is the prototypes in each class, which is denoted as $M=\{m_{ij}|i=1,...,C;\ j=1,...,K\}$. The parameters of $\theta$ and $M$ should be trained jointly in an end-to-end manner, and this can make them cooperate better with each other, which is beneficial for the performance of classification. To train the framework, we should first define the corresponding loss function. Besides, the loss function should be derivable with respect to $\theta$ and $M$ as well.  The loss function should also be closely related to the classification accuracy. In following subsections, we introduce multiple loss functions designed to train CPL.

\vspace{-0.3cm}\subsubsection{Minimum classification error loss (MCE)}
Minimum classification error (MCE) loss is firstly proposed by \cite{Juang1992discriminative}. We modify this loss function and make it applicable in our framework. In prototype learning, the discriminant function is defined as Eq. \ref{discriminant}. Then the misclassification measure of a sample from class $y$ is given by:
\begin{equation} \mu_{y}(x) = -g_{y}(x) + \left[\frac{1}{C-1}\sum_{j\neq y}g_{j}(x)^{\eta}\right]^{1/\eta} \end{equation}
when $\eta$ approaches infinity, the misclassification measure becomes:
\begin{equation} \mu_{y}(x) = -g_{y}(x) + g_{r}(x) \end{equation}
where $g_{r}(x)$ is the most competitive class, i.e.,
\begin{equation} g_{r}(x) = \max_{k\neq y}g_k(x)\end{equation}
then we can rewritten the misclassification measure as
\begin{equation} \mu_{y}(x)= \left\|f(x)-m_{yi}\right\|_{2}^{2} - \left\|f(x)-m_{rj}\right\|_{2}^{2} \end{equation}
where $m_{yi}$ is the closest prototype from the genuine class while $m_{rj}$ is the closest prototype from incorrect classes.
Then, the loss function is defined as:
\begin{equation} l((x,y);\theta,M) = \frac{1}{1+e^{-\xi\mu_{y}}}\label{mce}.\end{equation}
From the definition of the loss function, we can see that during minimizing the loss function, the $\mu_{y}$ is also minimized. This means $\left\|f(x)-m_{yi}\right\|_{2}^{2}$ is decreased and $\left\|f(x)-m_{rj}\right\|_{2}^{2}$ is increased, which acts like pull the feature closer to its class but push it away from the other classes. Thus, the training can help the framework to achieve better classification performance on the training samples.

The MCE loss is derivable with regard to $M$ and $f$, the derivatives can be calculated as:

\begin{equation} \frac{\partial l}{\partial f} = 2\xi l(1-l)(m_{rj}-m_{yi}). \end{equation}
\begin{equation} \frac{\partial l}{\partial m_{yi}} = 2\xi l(1-l)(m_{yi}-f(x)) \end{equation}
\begin{equation} \frac{\partial l}{\partial m_{rj}} = 2\xi l(1-l)(f(x) - m_{rj}) \end{equation}
and for the remainder prototypes, we have:
\begin{equation} \frac{\partial l}{\partial m} = 0. \end{equation}

Note that $f$ is the output of the CNN feature extractor, according to the error back propagation algorithm, the derivatives with regard to the parameters of CNN can be calculated start from $\partial l / \partial f$. Given the gradients of the loss function, the gradient based optimization methods can be used to update the whole framework.

\vspace{-0.3cm}\subsubsection{Margin based classification loss (MCL)}
For a training sample $(x,y)$, let $m_{yi}$ and $m_{rj}$ denote the closest prototypes from the correct class and the most competitive class respectively. If the sample is classified correctly, then $d(f(x),m_{yi}) < d(f(x),m_{rj})$ and we think the loss should be 0. If the sample is misclassified, then $d(f(x),m_{yi}) > d(f(x),m_{rj})$ and we think the loss exists in this situation. Naturally, the loss now can be defined as $d(f(x),m_{yi}) - d(f(x),m_{rj})$. Putting the two situations together, we can define the loss function as:
\begin{equation} l((x,y);\theta,M) = [d(f(x),m_{yi}) - d(f(x),m_{rj})]_{+}\label{mcl1} \end{equation}
To increase the classification ability of the framework, a margin is added to the loss function \ref{mcl1}, leading to the new margin based classification loss (MCL) function, which is denoted as:
\begin{equation} l((x,y);\theta,M) = [d(f(x),m_{yi}) - d(f(x),m_{rj})+m]_{+}\label{mcl2} \end{equation}
where $m$ is a positive number and acts as the margin. Compared with Eq. \ref{mcl1}, MCL is stricter, it penalize the framework even though it classifies some sample correctly (within the margin). Thus, it can increase the discriminative ability of the framework.

In order to apply MCL successfully, the margin $m$ should be carefully selected and it should have same scale with $d(f(x),m_{yi}) - d(f(x),m_{rj})$. However, the scale of $d(f(x),m_{yi}) - d(f(x),m_{rj})$ is unknown and we have to compute and estimate it from the training data. To avoid this problem, a generalized margin based classification loss (GMCL) function is proposed and defined as:
 \begin{equation} l((x,y);\theta,M) = \left[\frac{d(f(x),m_{yi}) - d(f(x),m_{rj})}{d(f(x),m_{yi}) + d(f(x),m_{rj})}+m\right]_{+}\label{mcl2} \end{equation}
In Eq. \ref{mcl2}, $-1 < \frac{d(f(x),m_{yi}) - d(f(x),m_{rj})}{d(f(x),m_{yi}) + d(f(x),m_{rj})} < 1$, thus we can simply choose margin $m$ from $(0,1)$.

Both MCL and GMCL are derivable with regard to $M$ and $f$, for MCL, the gradients can be computed by:
\begin{equation} \frac{\partial l}{\partial f} = \begin{cases} 2(m_{rj}-m_{yi}) & l>0 \\ 0 & l \le 0 \end{cases} \end{equation}
\begin{equation} \frac{\partial l}{\partial m_{yi}} = \begin{cases} 2(m_{yi}-f) & l>0 \\ 0 & l \le 0 \end{cases} \end{equation}
\begin{equation} \frac{\partial l}{\partial m_{rj}} = \begin{cases} 2(f-m_{rj}) & l>0 \\ 0 & l \le 0 \end{cases} \end{equation}
for other prototypes, we have
\begin{equation}\partial l / \partial m_{ij} = 0\end{equation}
Similarly, the gradients of GMCL can also be computed directly and we do not list the equations any more.

Equally, the gradients of the loss function with regard to $\theta$ can also be computed by the error back propagation algorithm beginning from $\partial l / \partial f$. In MCL and GMCL, the framework is updated only when the loss exists and during the updating, only two prototypes are trained but the other prototypes are kept unchanged.

\vspace{-0.3cm}\subsubsection{Distance based cross entropy loss (DCE)}
In our CPL framework, the distance can be used to measure the similarity between the samples and the prototypes. Thus, the probability of a sample $(x,y)$ belonging to the prototype $m_{ij}$ can be measured by the distance between them:
\begin{equation} p(x\in m_{ij}|x)\propto -\left\|f(x)-m_{ij}\right\|_{2}^{2}.\end{equation}
To satisfy the non-negative and sum-to-one properties of the probability, we further define the probability $p(x\in m_{ij}|x)$ as:
\begin{equation} p(x\in m_{ij}|x) = \frac{e^{-\gamma d(f(x),m_{ij})}}{\sum_{k=1}^{C}\sum_{l=1}^{K}e^{-\gamma d(f(x),m_{kl})}}\label{prototypeprob} \end{equation}
where $d(f(x), m_{ij}) = \left\|f(x)-m_{ij}\right\|_{2}^{2}$ represents the distance between $f(x)$ and $m_{ij}$. $\gamma$ is a hyper-parameter that control the hardness of probability assignment. Given the definition of $p(x\in m_{ij}|x)$, we can further define the probability of $p(y|x)$ as:
\begin{equation}p(y|x)=\sum_{j=1}^{K}p(x\in m_{yj}|x)\label{classprob}.\end{equation}
Based on the probability of $p(y|x)$, we can define the cross entropy (CE) loss under our framework as:
\begin{equation} l((x,y);\theta,M) = -logp(y|x)\label{dce} .\end{equation}
This loss function is defined based on the distance, to distinguish it from the traditional cross entropy loss, we call it distance based cross entropy (DCE) loss. From Eq. \ref{prototypeprob}, \ref{classprob} and \ref{dce}, we can see that minimizing the loss function essentially means decreasing the distance between the samples with the prototypes which come from the genuine class of the samples.

Obviously, the DCE is also derivable with regard to $M$ and $f$, here we don't list the equation of the corresponding gradients any more.
Similarly, the derivatives with regard to the parameters of CNN can also be calculated starting from $\partial l / \partial f$ by the chain rule. Compared with MCL and GMCL, DCE updates all the prototypes every time during the training, thus it converges faster than the MCL and GMCL.

\subsection{Generalized CPL with prototype loss}\label{sect:pl}
The loss functions defined in section \ref{training} are used as measurements of classification accuracy, and by minimizing these losses, we can train the model to classify the data correctly. However, directly minimizing the classification loss may lead to over-fitting. In light of this, we propose a new prototype loss (PL) as a regularization, which acts like a generative model to improve the generalization performance of CPL.

From the prediction function defined in Eq. \ref{prediction}, we can derive the decision boundary of the CPL:
\begin{equation} \left\|f-m_{ij}\right\|_{2}^{2}= \left\|f-m_{kl}\right\|_{2}^{2}\end{equation}
\begin{equation} 2f\cdot (m_{kl} - m_{ij}) + \left\|m_{ij}\right\|_{2}^{2} - \left\|m_{kl}\right\|_{2}^{2}=0.\end{equation}
We can see that the resulted decision boundary is still linear. Like traditional CNN framework for classification, the CPL still separates the whole feature space and the learned representation is still linearly separable. As stated before, this kind of framework is not robust, it can not reject the samples from unseen classes and can not be extended to new classes conveniently. To overcome this problem, a new loss function called prototype loss (PL) is added in our framework, which is defined as:
\begin{equation} pl((x,y);\theta, M)=\left\|f(x)-m_{yj}\right\|_{2}^{2} \label{pl} \end{equation}
where $m_{yj}$ is the closest prototype with $f(x)$ from the corresponding class $y$. The prototype loss can be combined with the classification loss defined in section \ref{training} to train the model. Then the total loss can be defined as:
\begin{equation} loss((x,y);\theta,M) = l((x,y);\theta,M) + \lambda pl((x,y);\theta,M) \label{totalloss} \end{equation}
where $\lambda$ is a hyper-parameter which control the weight of prototype loss. The PL can also be viewed as the maximum-likelihood (ML) regularization \cite{liu2004effects, liu2004discriminative} which is widely used in pattern recognition \cite{jin2010regularized, liu2010one}.

PL can further boost the performance of CPL, because: (1) PL pull the features of samples close to their corresponding prototypes, making the features within the same class more compact, this can implicitly increase the distance between the classes, which is beneficial for classification; (2) the classification loss stresses the separation property of the representation and the prototype loss stresses the compactness property of the representation, by combining them together, we can learn intra-class compact and inter-class separable representations, which are more robust and more appropriate for rejection and open set problems. We denote CPL equipped with PL as generalized convolutional prototype learning (GCPL) \footnote{The main codes of CPL and GCPL can be found at https://github.com/YangHM/Convolutional-Prototype-Learning}.

\vspace{-0.1cm}\section{Application of GCPL}\label{sect:application}
Besides classification, GCPL can also be used for rejection and class-incremental learning. In this paper, we did not invent new rejection and class-incremental learning methods, but only show that our framework is suitable for these two tasks. Most rejection strategies are based on the probabilities (confidences) produced by the softmax layer of CNN model. In our framework, we can also obtain classification probabilities by Eq. \ref{classprob}, so the same strategies can also be used in our framework. The distance is also a meaningful measurement for the classification confidence in our framework, thus the same strategies can also be implemented based on the distance outputted by GCPL.

For class-incremental learning, we only consider the most general case. That is, given a trained framework and some samples from a new class, we should extend the framework to recognize the new class correctly and keep the accuracy on the old classes. In GCPL framework, the learned representations have better clustering property, for samples of the new class, the resulted features are also compact. Thus we can cluster the features of new class or use the mean of these features as the prototypes for the new class. Therefore, the framework can be expanded to the new class easily. In the following experimental section, we will show that this approach is very effective for class-incremental learning.

\section{Experiments}

\subsection{Experiments and analysis on MNIST \cite{lecun1998gradient}}\label{sect:mnistexperiment}

\begin{figure}
\begin{center}
\includegraphics[width=0.8\linewidth]{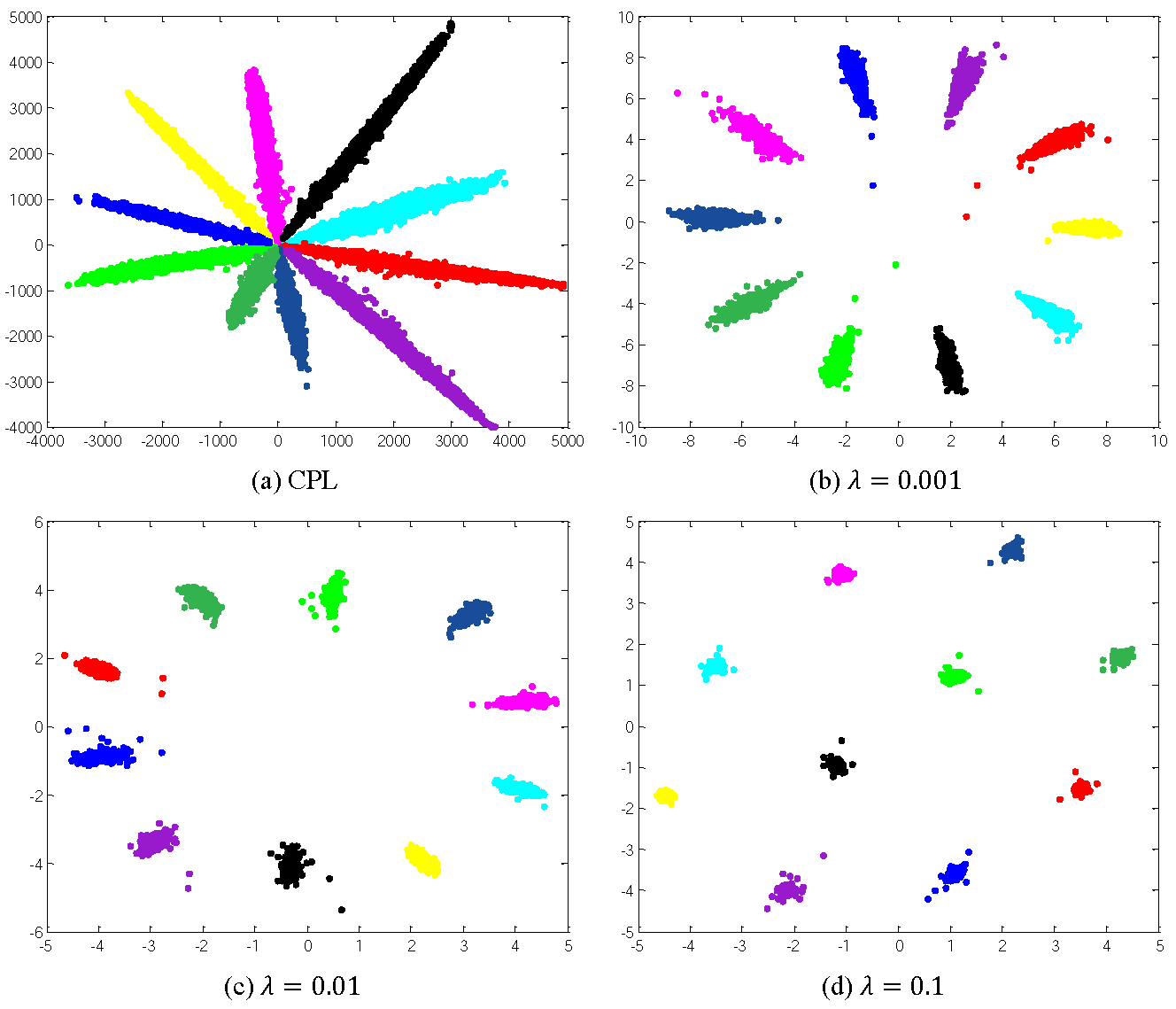}
\end{center}
   \caption{The learned representations of CPL and GCPL on MNIST. Different colors represent different classes}
\label{fig:visual}
\vspace{-0.5cm}
\end{figure}

In this experiment, the architecture of CNN feature extractor is the same as the network used in \cite{Wen2016A} with the ReLU activation function. The output of the CNN feature vector is set to be two, thus we can directly plot the features on the 2-D surface for visualization. We maintain one prototype in each class and train the framework under the DCE loss and DCE+PL loss respectively. Meanwhile, we also give the accuracy of the traditional softmax based CNN framework under the same architecture as \cite{Wen2016A}. We set the initial learning rate as 0.001, the batch size as 50, and the hyper-parameter $\gamma$ in DCE as 1.0 during the training. The final results can be seen in table \ref{mnistacc} and Fig. \ref{fig:visual}.

\begin{table}
\begin{center}
\begin{tabular}{|c|c|c|}
\hline
\multicolumn{2}{|c|}{method} & test accuracy (\%) \\
\hline
\multicolumn{2}{|c|}{soft-max} & 99.08 \\
\hline
\multicolumn{2}{|c|}{CPL (DCE)} & 99.28 \\
\hline
\multirow{3}{*}{GCPL(DCE+PL)} & $\lambda=0.001$ & 99.33  \\
\cline{2-3}
& $\lambda=0.01$ & 99.29 \\
\cline{2-3}
& $\lambda=0.1$ & 99.30 \\
\hline
\end{tabular}
\end{center}
\caption{Test accuracy of different methods on MNIST}\label{mnistacc}
\vspace{-0.1cm}
\end{table}

From table \ref{mnistacc}, we can see that our new proposed CPL framework can achieve comparable performance with traditional softmax based CNN under the same structure. Moreover, by cooperating with PL, our GCPL can achieve even better results. This demonstrates that the PL, which acts as an implicit regularization, is beneficial for classification as well. Meanwhile, we can also see that the test accuracy is not sensitive to the parameter $\lambda$, the change of $\lambda$ did not greatly impact the results. From Fig. \ref{fig:visual} (a), we can see that when only use the classification loss, the resulted representations are still linear separable, this demonstrates that our analysis in section \ref{sect:pl} is correct. From Fig. \ref{fig:visual} (b) to (d), we can see the learned representations under GCPL framework are really inter-class separable and intra-class compact. With the increasing of the weight $\lambda$ on PL loss, the learned representations within the same class become more and more compact. This demonstrate our GCPL framework can really learn robust and discriminative representations.

\subsection{Experiments and analysis on CIFAR-10}\label{sect:cifar}
We realized several CNN structures on CIFAR-10 \cite{krizhevsky2009learning}, including the model C appeared in \cite{Springenberg2014Striving}, a modified version of model C that adds batch normalization layers after each convolutional and fully connected layers in model C, residual net 20 and residual net 32 in \cite{He2016Deep}. Then we compare the performance of the traditional softmax based classification, CPL, and GCPL methods under these CNN structures. In all experiments, we maintain one prototype for each class and the prototypes are initialized as zero vectors. The images data are whitened and contrast normalized following \cite{goodfellow2013maxout}. The experimental results are shown in table~\ref{table:cifaracc}.

\begin{table}
\begin{center}
\begin{tabular}{|c|c|c|c|}
\hline
CNN structure & soft-max & CPL & GCPL \\
\hline
model C \cite{Springenberg2014Striving} & 90.26 \cite{Springenberg2014Striving} & 90.70 & 90.80  \\
model C with BN & 91.37 & 91.59 & 91.90 \\
ResNet 20 & 91.32 & 91.46 & 91.63 \\
ResNet 32 & 92.50 & 92.60 & 92.63 \\
\hline
\end{tabular}
\end{center}
\caption{The accuracy of different CNN structures and different models on CIFAR-10}\label{table:cifaracc}
\vspace{-0.1cm}
\end{table}

From table \ref{table:cifaracc}, we can see that the proposed CPL framework can achieve comparable or even better results than the traditional softmax based classification method under different network structures. This further demonstrates the efficiency and generality of the proposed framework. Besides, the GCPL framework, which has an additional prototype loss during training, performs best in all experiments on CIFAR-10.

\subsection{Experiments and analysis on OLHWDB}
Online handwriting database (OLHWDB \cite{liu2011casia}) is a large scale Chinese handwriting dataset. Following the settings in \cite{Zhang2016Online} and \cite{yin2013icdar}, the training and test datasets include 2,697,673 and 224,590 samples respectively, which are come from 3755 classes. We use the same CNN structure as \cite{Zhang2016Online} and make little modifications with batch normalization and ReLU to improve training process. We mainly test the GCPL framework on this dataset and adopt different classification loss functions to study their influence on the performance. We maintain only one prototype for each class and the prototypes are initialized as the mean of training features in the corresponding classes. We use the same data pre-processing method as \cite{Zhang2016Online}. The experimental results are listed in table \ref{table:olhwdbacc}. From table \ref{table:olhwdbacc}, we can see that our framework also performs well or even better on large scale classification problems, which again demonstrate its efficiency and generality.

\begin{table}
\begin{center}
\begin{tabular}{|c|c|}
\hline
loss function & accuracy (\%) \\
\hline
soft-max & 97.55 \cite{Zhang2016Online} \\
\hline
MCE & 97.35 \\
MCL & 97.61 \\
GMCL & 97.36 \\
DCE & 97.58 \\
\hline
\end{tabular}
\end{center}
\caption{The accuracy of GCPL on OLHWDB dataset}\label{table:olhwdbacc}
\vspace{-0.5cm}
\end{table}

Note that our purpose is not to achieve significantly better accuracy than previous softmax-based CNN model, we only want to show that, from the accuracy perspective, our framework can match or work slightly better than traditional CNN model. The advantage of our model in another perspective is that it can significantly improve the robustness of pattern recognition. In following subsections, we will show this from the viewpoints of rejection and class-incremental learning.

\subsection{Experiments for rejection}
To test the robustness of our GCPL framework, we further conduct experiments for rejection. We firstly train a network on MNIST training set. To evaluate its robustness, we use two test sets (MNIST and CIFAR-10 test sets) for this network. The CIFAR-10 test samples are not digits, and therefore, they should be viewed as outliers and then be rejected by this network. At the same time, the samples from the MNIST test set should still be accepted since they are from the same domain as the training data. Actually, the rejection and acceptance performance are closely coupled, we can only get a tradeoff between them.

To fairly evaluate the performance, we use two measurements of acceptance rate (AR) and rejection rate (RR). AR denotes the percentage of accepted samples in MNIST test set (how many MNIST samples have been accepted), while RR denotes the percentage of rejected samples in CIFAR-10 test set (how many CIFAR-10 samples have been rejected). The higher of these two measurements, the better of the model in robustness. We adopt the most frequently used threshold-based rejection strategy, i.e., if the output confidence for a sample is larger than the pre-defined threshold, then it will be accepted, otherwise it will be rejected. The confidence can be obtained by the output probability or distance in GCPL (section \ref{sect:application}). Different from the probability, the smaller distance represents larger confidence. We use the same structure of CNN feature extractor as section \ref{sect:mnistexperiment}. For comparison, we also test the rejection performance of traditional softmax based framework under the same CNN structure. The results are showed in table \ref{table:rejection}. Note these results are obtained by using different (smoothly changed) thresholds to give the AR-RR tradeoffs.

\begin{table}\
\begin{center}
\begin{tabular}{|c|c|c|c|c|c|}
\hline
 \multicolumn{2}{|c|}{softmax} & \multicolumn{2}{|c|}{GCPL Prob} &\multicolumn{2}{|c|}{GCPL Dist}\\
\hline
AR & RR & AR & RR & AR & RR\\
\hline
100.0 & 0.000    & 100.0 & 0.000        & \textbf{94.20} & \textbf{100.0}\\
99.98 & 0.200    & 99.99 & 15.67  & \textbf{97.39} & \textbf{100.0}\\
99.72 & 8.110    & 99.80 & 46.12  & \textbf{98.07} & \textbf{100.0}\\
99.14 & 25.17    & 99.39 & 87.07  & \textbf{98.43} & \textbf{100.0}\\
98.52 & 40.60    & 99.30 & 93.43  & \textbf{98.57} & \textbf{99.99}\\
97.61 & 57.54    & 99.21 & 96.31  & \textbf{98.73} & \textbf{99.99}\\
83.95 & 71.66    & 98.96 & 98.69  & \textbf{98.89} & \textbf{99.99}\\
76.67 & 85.97    & 98.73 & 99.46  & \textbf{99.09} & \textbf{99.99}\\
75.49 & 98.02    & 98.21 & 99.86  & \textbf{99.20} & \textbf{99.99}\\
\hline
\end{tabular}
\end{center}
\caption{The tradeoff between acceptance rate AR (\%) and rejection rate RR (\%) for different methods.}\label{table:rejection}
\vspace{-0.5cm}
\end{table}

From table \ref{table:rejection}, we can see that the softmax-based model is confused by the MNIST and CIFAR-10 samples, the high AR and high RR is not able to coexist. This explains that the softmax-based model is not robust in outlier detection. Differently, our GCPL model can achieve better rejection performance and meantime keep satisfactory acceptance rate. For example, while over 99\% CIFAR-10 samples being rejected, we can still keep over 99\% MNIST samples accepted. This is a significant advantage compared with softmax based approach, which demonstrates the robustness of the GCPL framework.

\subsection{Experiments for class-incremental learning}
\begin{figure}[h]
\begin{center}
\includegraphics[width=0.99\linewidth]{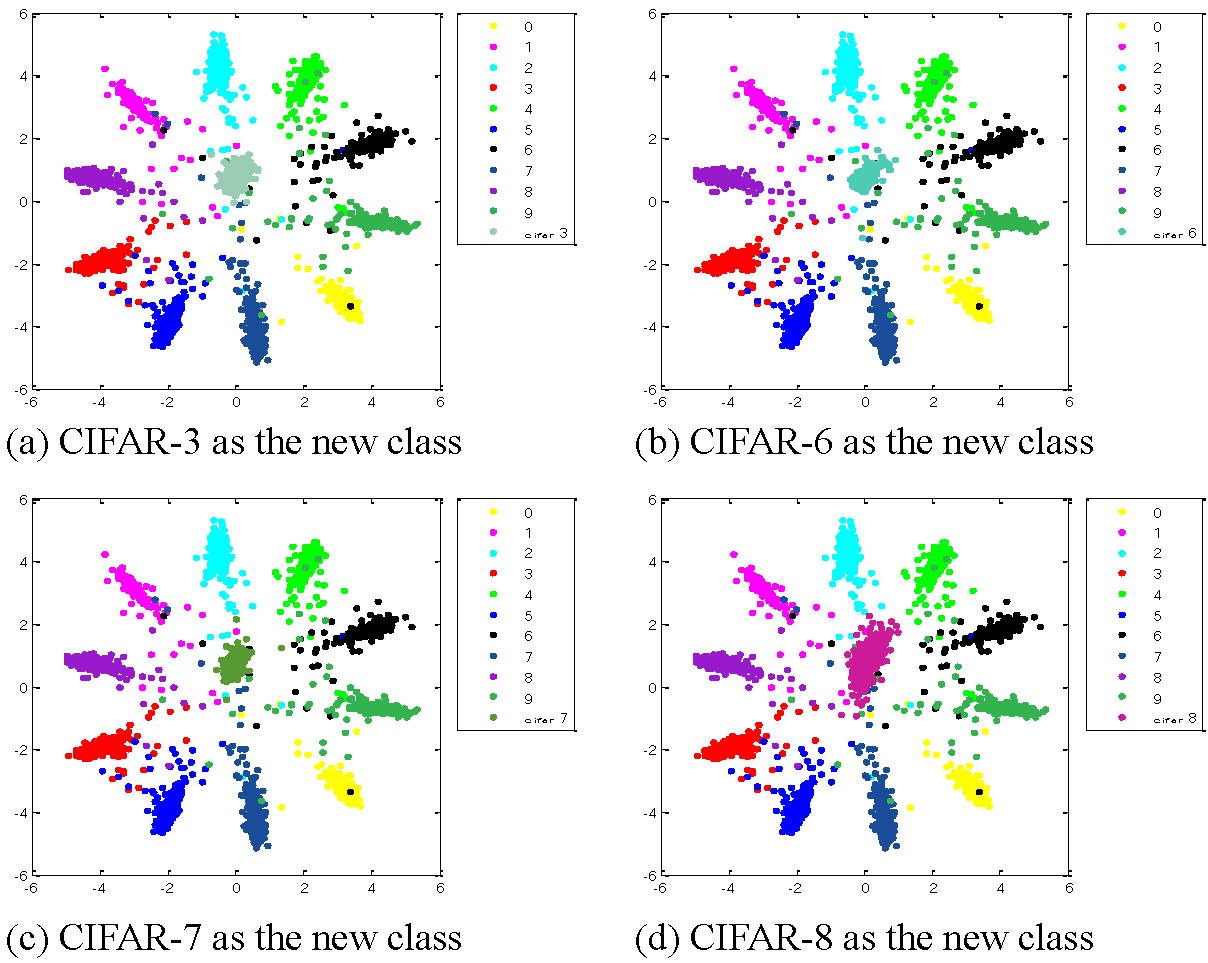}
\end{center}
   \caption{The learned representations for both known and unknown (new) classes}
\label{fig:incremental}
\vspace{-0.2cm}
\end{figure}

We conduct experiments on MNIST and CIFAR dataset to demonstrate the superiority of GCPL framework for class-incremental learning. We treat all MNIST samples (from class 0 to 9) as the known class data and choose one class from CIFAR-10 as the new class (which should be learned incrementally). We use the same network structure as described in section \ref{sect:mnistexperiment} and train the GCPL on the MNIST training set, then we feed the test data from both known and unknown classes to the trained GCPL and obtain their representations, the results are shown in Fig. \ref{fig:incremental}.

From Fig. \ref{fig:incremental}, we can find that the learned representations for both known and unknown (new) classes are intra-class compact and inter-class separable, this again demonstrates the robustness of the GCPL framework. Based on such representations, we can simply use the mean of the training data from the new class as the prototype for the new class, then we can directly extend the GCPL to make predictions for both the new class and the previous classes. To evaluate the accuracy of GCPL for class-incremental learning, we further give the accuracy of the extended GCPL on the test samples from both known and new classes (total 11 classes) in table \ref{table:acc}. From table \ref{table:acc}, we can see the GCPL still keeps high performance when extended to the new class. Note that in this class-incremental learning process, we did not re-train any part of the model, and due to the advantage of prototype-based decision making, we can directly add a new prototype to represent the new class. This demonstrates the advantage of GCLP for class-incremental learning.

\begin{table}
\begin{center}
\begin{tabular}{|c|c|}
\hline
new class ID (CIFAR) & test accuracy (11-class) \\
\hline
0 & 99.23 \\
1 & 99.23 \\
3 & 99.24 \\
5 & 99.23 \\
6 & 99.24 \\
7 & 99.22 \\
8 & 99.20 \\
9 & 99.23 \\
\hline
without new class & 99.27 \\
\hline
\end{tabular}
\end{center}
\caption{Accuracy (\%) of GCPL on test samples from both known and new classes}\label{table:acc}
\vspace{-0.39cm}
\end{table}

\subsection{Experiments with small sample size}
Another aspect of robust classification is the ability to deal with small sample size (SSS). To further demonstrate the robustness of GCPL, we conduct experiments under the condition of SSS. We use different number (percentages) of samples in MNIST training set to train the models and then observe their accuracy on the MNIST test set. For GCPL, We use the same net structure and training settings as described in section \ref{sect:mnistexperiment}, the weight $\lambda$ for prototype loss is 0.001. For comparison, we also give the test accuracy of the traditional softmax-based CNN framework under the same architecture and training data. We repeat the experiments for five times and show the statistical results in table~\ref{table:small}.

\begin{table}[h]
\begin{center}
\begin{tabular}{|c|c|c|}
\hline
sample size (\%) & soft-max & GCPL \\
\hline
100 & $99.08\pm0.10$ & $99.33\pm0.10$ \\
50 & $98.07\pm0.39$ & $99.12\pm0.10$ \\
30 & $92.68\pm4.52$ & $98.89\pm0.10$ \\
10 & $86.12\pm6.00$ & $97.80\pm0.22$ \\
5 & $73.95\pm6.10$ & $96.44\pm0.40$ \\
3 & $50.79\pm17.44$ & $94.90\pm0.58$ \\
\hline
\end{tabular}
\end{center}
\caption{Test accuracy (\%) under different percentages of training samples.}\label{table:small}
\vspace{-0.3cm}
\end{table}

From table~\ref{table:small}, we can find that the decreasing of training samples has less impact on the performance of GCPL. Compared with the softmax-based framework, the performance of GCPL declines much slower when the sample size is reduced. In particular, when training with very small number of training samples (e.g., with only 5\% or 3\% training samples), the test accuracies for softmax-based framework are very low (with large variance), which demonstrates that the softmax-based model is not robust for SSS problem. On the contrary, under the same situation, the GCPL is still very effective for different sizes of training samples. GCPL not only achieves much higher accuracy but also shows more stable results with smaller variances. This again demonstrates the robustness of GCPL in dealing with SSS problem.

\subsection{Experiments with multiple prototypes}
In all previous experiments, we set the number of prototypes $K$ in each class as 1. In this section, we adopt different values for $K$ and investigate its effect on the classification performance. We use model C (described in section \ref{sect:cifar}) with DCE loss, and conduct the experiment on CIFAR-10 dataset. For different $K$, we use the same settings for the hyper-parameters during the training, the results are shown in table \ref{table:multi}.

\begin{table}[h]
\begin{center}
\begin{tabular}{|c|c|c|c|c|c|}
\hline
$K$ & 1 & 2 & 3 & 4 & 5 \\
\hline
accuracy (\%) & 90.70 & 90.40 & 90.37 & 90.67 & 90.46 \\
\hline
\end{tabular}
\end{center}
\caption{Test accuracy (\%) under different values of $K$.}\label{table:multi}
\vspace{-0.3cm}
\end{table}

From table \ref{table:multi}, we can see that more prototypes didn't lead to better results. Actually, CNN is very powerful for feature extraction, even though the initial intra-class distribution may be very complex, after CNN transformation, it can still be well modeled with a single Gaussian distribution (a single prototype). However, in more complicated scenarios, where the data distributions are difficult to model, more prototypes may be beneficial.

\section{Conclusion}
This paper proposed a convolutional prototype learning (CPL) framework for pattern recognition. Different from the softmax-based models, CPL directly learn multiple prototypes (in convolutional feature space) for each class and then use prototype matching for decision making. CPL can achieve comparable or even better classification accuracy than softmax-based CNN models. To further improve the robustness, we propose a prototype loss (PL) to increase the intra-class compactness, resulting in a generalized CPL (GCPL) model. The GCPL has great advantage compared with traditional CNN models in the perspectives of outlier rejection and class-incremental learning. In future, we will try to conduct more detailed experiments to evaluate other properties of GCPL such as the adaptation ability in changing environment, dealing with weakly labeled data, and so on.

\section*{Acknowledgments}
This work has been supported by the National Natural Science Foundation of China (NSFC) Grants 61721004 and 61633021, and NVIDIA NVAIL program.


{\small
\bibliographystyle{ieee}
\bibliography{ref}
}

\end{document}